\pdfoutput=1

\documentclass[11pt]{article}

\usepackage[final]{acl}

\usepackage{times}
\usepackage{latexsym}

\usepackage[T1]{fontenc}

\usepackage[utf8]{inputenc}

\usepackage{microtype}

\usepackage{inconsolata}
\usepackage{enumitem}
\usepackage{graphicx}
\usepackage{multirow}
\usepackage{makecell}  

\usepackage{colortbl}
\usepackage{multirow}
\usepackage{xcolor}

\definecolor{basicLight}{RGB}{230, 242, 255} 
\definecolor{basicDark}{RGB}{179, 217, 255} 

\definecolor{preExploreLight}{RGB}{255, 235, 230} 
\definecolor{preExploreDark}{RGB}{255, 194, 179} 

\definecolor{inventoryLight}{RGB}{230, 255, 230} 
\definecolor{inventoryDark}{RGB}{179, 242, 179} 

\definecolor{syntheticLight}{RGB}{242, 230, 255} 
\definecolor{syntheticMedium}{RGB}{217, 179, 255} 
\definecolor{syntheticDark}{RGB}{191, 128, 255} 

\title{\textit{Plant in Cupboard, Orange on Rably, Inat Aphone}.\\Benchmarking Incremental Learning of Situation and Language Model using a Text-Simulated Situated Environment}

\author{
 \textbf{Jonathan Jordan\textsuperscript{1}},
 \textbf{Sherzod Hakimov\textsuperscript{1}},
 \textbf{David Schlangen\textsuperscript{1,2}}
\\
\\
 \textsuperscript{1}Computational Linguistics, Department of Linguistics\\University of Potsdam, Germany\\
 \textsuperscript{2}German Research Center for Artificial Intelligence (DFKI), Berlin, Germany
\\
\texttt{firstname.lastname@uni-potsdam.de} 
}

\begin{document}
\maketitle

\begin{abstract}
Large Language Models (LLMs) serve not only as chatbots but as key components in agent systems, where their common-sense knowledge significantly impacts performance as language-based planners for situated or embodied action. We assess LLMs' incremental learning (based on feedback from the environment),  and controlled in-context learning abilities using a text-based environment. We introduce challenging yet interesting set of experiments to test i) how agents can incrementally solve tasks related to every day objects in typical rooms in a house where each of them are discovered by interacting within the environment, ii) controlled in-context learning abilities and efficiency of agents by providing short info about locations of objects and rooms to check how faster the task can be solved, and finally iii) using synthetic pseudo-English words to gauge how well LLMs are at inferring meaning of unknown words from environmental feedback. Results show that larger commercial models have a substantial gap in performance compared to open-weight but almost all models struggle with the synthetic words experiments.

\end{abstract}

\section{Introduction}

\begin{figure}[ht!]
    \centering
    \includegraphics[width=1.0\linewidth]{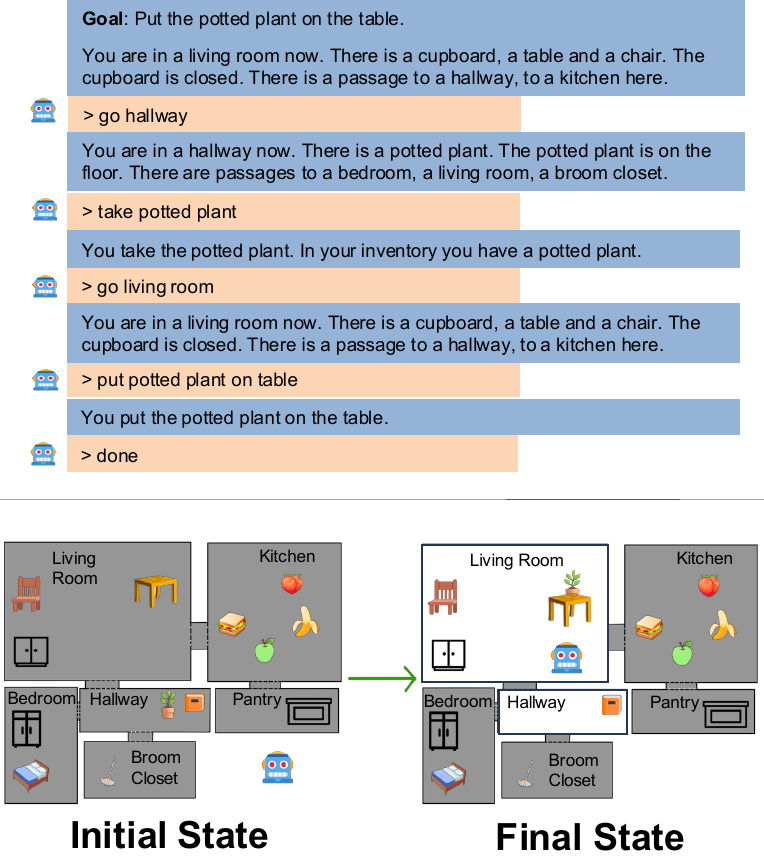}
    \caption{Sample representation of an episode in our game. The agent \protect\includegraphics[width=0.023\textwidth]{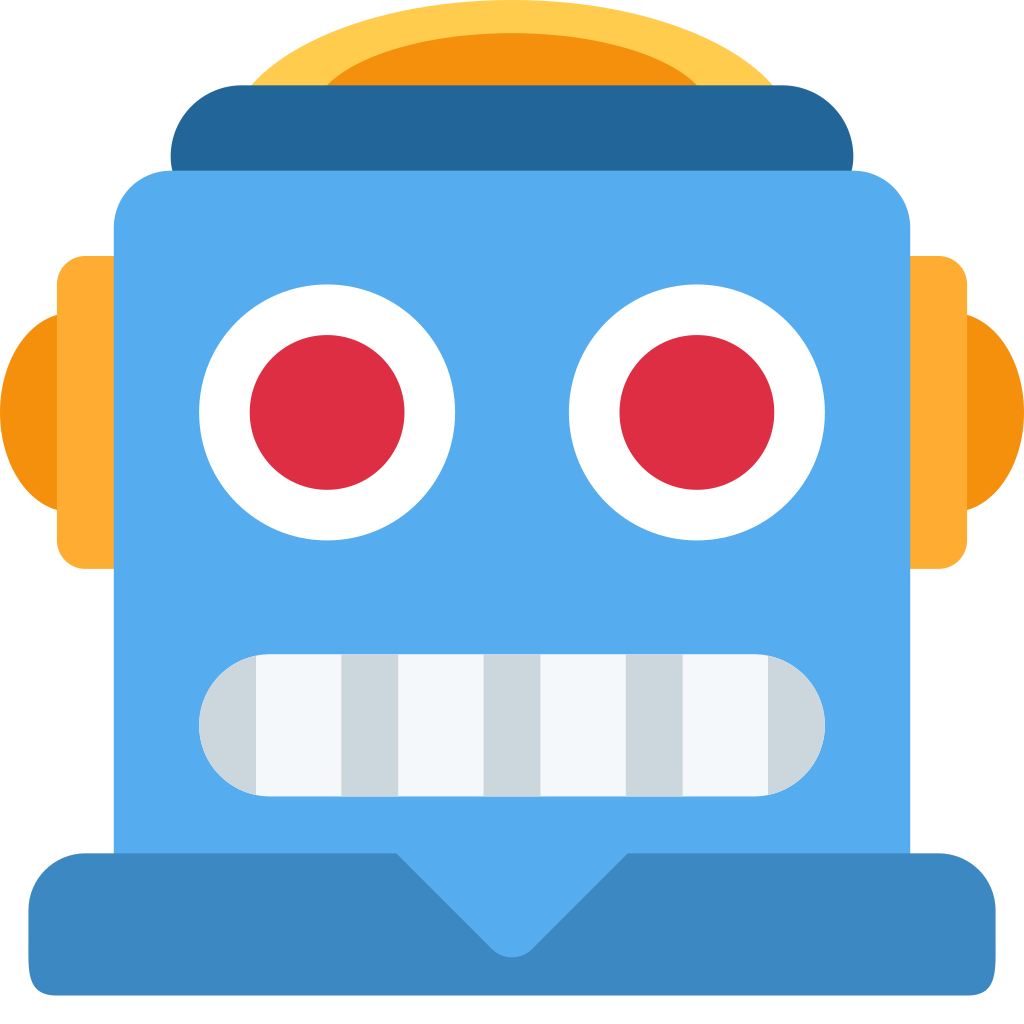} (an LLM) is given a task and randomly assigned to some room (living room). The environment provides feedback for every action (go, take, put, etc.) of an agent. The top part is how the game is played in a textual world. The bottom part is the visual representation of initial and final states, also indicating the knowledge gained by the agent.}
    \label{fig:intro-ex}
\end{figure}

Theoretical reasoning, which involves deriving factual conclusions from given premises, has been extensively studied in the context of large language models (LLMs)~\cite{DBLP:conf/nips/BrownMRSKDNSSAA20,DBLP:conf/nips/Wei0SBIXCLZ22,DBLP:conf/nips/KojimaGRMI22}. There has been less work focusing on incremental learning \cite{Schlangen-2023} of agents with regards to how easily they generalise to unknown environments. While theoretical reasoning can be evaluated using reference propositions, incremental learning presents unique challenges because the agent is required to observe consequences of actions in an environment (can be unfamiliar one) and learn how to interact step-by-step. There is significant interest in using LLMs for embodied AI tasks in robotics and simulations.~\cite{DBLP:conf/corl/IchterBCFHHHIIJ22}. Many of the existing benchmarks ( WebArena~\cite{zhou2024webarenarealisticwebenvironment}, ALFRED~\cite{alfworld}, AI2-THOR~\cite{kolve2022ai2thorinteractive3denvironment}, Balrog Arena~\cite{paglieri2024balrog}) include complex environments or require computational resources to execute vision-related tasks. In contrast, text-based interactive fiction (IF) environments such as TALES~\cite{cui2025tales} or TextWorld~\citep{textworld}, where a situation is described textually and updated in response to textual commands, make it possible to study reasoning abilities that go from goal and situation description to goal-directed action in greater isolation.

In this paper, we focus on incremental learning capabilities of agents (LLMs) under controlled and uncontrolled conditions. Agents have to uncover the restrictions and rules of the task by interacting and taking the feedback (in-context learning) by environment into account. We also examine to what degree the underlying common-sense knowledge embedded in LLMs helps to navigate the tasks and how much of it is due to their generalisation capabilities. We expand this by introducing our text-based \textit{AdventureGame}\footnote{Source code: \url{https://github.com/clembench/clembench/tree/main/adventuregame} (inspired by the ``Colossal Cave Adventure'' game by W.~Crowther: \url{https://ifdb.org/viewgame?id=fft6pu91j85y4acv})} environment to control the applicable level of common-sense world knowledge, for testing situated language environment understanding, spatial navigation, and instruction following abilities. We introduce various experiments to assess the effect of common-sense knowledge and generalisation capabilities where the agent (LLM acting on an environment) has to perform \textit{home delivery} tasks with common objects within room types or has to infer the meaning of synthetic words by solely interacting and getting feedback from the environment. Figure~\ref{fig:intro-ex} (top)
shows an example of an interaction, while the bottom part is the visual representation of what has changed going from an initial state to the final one.
 Our contributions are as follows: 
\begin{itemize}[itemsep=0pt, parsep=0pt]
    \item A set of experiments based on an established object delivery task in a typical home setting, with varied levels of complexity, situated interaction task demands and information to be acquired.
    \item Synthetic words experiment to assess reliance on common-sense knowledge and incremental learning capabilities
    \item Quantitative and qualitative comparison of recent large language models (LLM) of varying sizes
    \item In-depth analysis to assess models' behaviour, examining good interactions and common failure cases. 
\end{itemize}

\begin{figure*}[h]
    \centering
    \includegraphics[width=\textwidth]{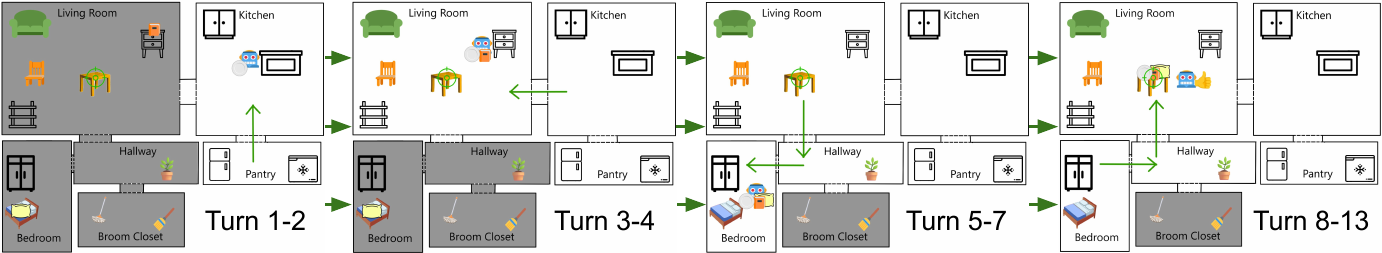}
    \caption{Simplified illustration of \textit{AdventureGame} interaction. The agent \protect\includegraphics[width=0.023\textwidth]{figures/robot.png} is controlled by an LLM. The task is \textit{put the plate, book and pillow on the table} (marked by green crosshair). The agent starts in the pantry. Unexplored rooms are gray. The agent first goes to the kitchen, \textit{takes the plate from the counter} (Turn 1-2), then \textit{goes to the living room, takes the book} (Turn 3-4). Next, it \textit{goes to the hallway}, and from there \textit{to the bedroom where it picks up the pillow} (Turn 5-7), then it \textit{returns to the living room}, \textit{puts the carried objects on the table} (Turn 8-13).}
    \label{fig:simple-illu}
\end{figure*}

\section{Related Work}

\textbf{In-context Learning from Feedback}: LLMs acquire situational information in IF environments through self-guided in-context learning (ICL) from observations and failure feedback. Testing LLM generalisability requires test instances unlikely to be seen during training to avoid reliance on memorized common-sense knowledge. \textit{BlocksWorld} obfuscates block-manipulation words~\cite{gragera2023exploring, valmeekam2024llms, plan-bench}, while \citet{eisenschlos2023winodict} tested ICL by providing pseudo-English words with explanations to measure proper application in WinoGrad schema. \citet{reflexion} use ``verbal reinforcement learning'' with contextual feedback for environment interaction without weight training. \citet{shi2024why} found that smaller models rely on semantic priors while larger ones override mappings based on input context in label-flipping tasks. \textit{AdventureGame} extends IF benchmarking with synthesized words in household tasks, from partial to fully synthetic non-function vocabulary.

\textbf{Interactive Fiction (IF) Environments}: These environments have been used to \textit{train and compare agents} where they combine a specific set of agentic challenges: partial observability, large world state and action space, requiring exploration and common-sense reasoning, in addition to the text-only interaction and feedback. It has been shown that performance in text-only IF environments is transferable to embodied environments (\citet{alfworld}, \citet{jansen2021systematic}). 
Interactive fiction (IF) environments have also been used to \textit{compare LLM performance} in regards to their world modelling, task solving and planning capabilities (\citet{wang2022scienceworld}, \citet{tan2023text}, \citet{tsai2023can}, \citet{ma2024agentboard}, \citet{gioacchini2024agentquest}).
Jericho~\cite{hausknecht2020interactive} provides a framework for classic IF games, which remain challenging for LLMs due to complexity and humorously nonsensical solutions. \textit{AdventureGame} condenses core IF challenges while avoiding overly difficult genre aspects. TextWorld~\cite{textworld} extends Jericho for large-scale simplified IF generation and agent benchmarking. \textit{AdventureGame} enables similar generation but records detailed interaction scores and exploration data. ALFWorld~\cite{alfworld} combines 3D-embodied simulation with text-only pre-training, demonstrating text interaction transferability to multimodal domains. TALES~\cite{cui2025tales} benchmarks LLM reasoning by combining the above IF environments with \textit{Simon Says} and ScienceWorld~\cite{wang2022scienceworld}.

\textit{AdventureGame} has a unique focus on text-only household object placement tasks by requiring the tested LLMs to perform in-context learning from minimal natural language feedback to assess the core reasoning, instruction following, situated action selection, and, most importantly, the effects of synthetic vocabulary to test generalization beyond training-embedded common-sense knowledge.

\section{\textit{AdventureGame}: IF Environment}

In this section, we describe the details of the game. A simplified illustration is given in Figure~\ref{fig:simple-illu}, where an LLM controls an agent to perform the given task. We implemented the game using the clembench framework~\cite{clembench2023} (prompts are provided in Appendix~\ref{sec:initial-prompts}).

\subsection{Task Definition}
Interactive Fiction games can be considered partially observable Markov Decision processes~\cite{KAELBLING199899}, with the player having only partial information about the game's world state from textual observations. The player must first discover all task-relevant locations and objects, making exploration as much a part of solution strategies as effectively executing a plan for manipulating objects to reach the game's goals. To do so successfully, the player has to model environment states (like room connections and locations of objects), as well as the large action space and state transitions resulting from actions, especially over multiple turns.

The task is defined by the tuple $\langle G,S,A,O,T\rangle$ with a set of goal states $G$, state space $S$, valid action space $A$, observation space $O$ (including IF interpreter feedback), and transition function $T : S\times A \rightarrow S$.

An agent with policy $\pi$ makes predictions at time step $t$ based on goals in $G$ and memory $m_t = \{o_j, a_j, o_{j+1}, a_{j+1}, . . . o_t\}, 0 \leq j < t$, which is a sequence of actions and observations. This agent trajectory $\tau = [s_0, a_0, s_1, a_1, . . . s_t]$ is formulated by policy and environmental state transitions as below where a time step $t$ is one turn of the game:
$$p_\pi(\tau) = p(s_0) \prod^{T}_{t=0}\pi(a_t|G, s_t, m_t)\tau (s_{t+1}|s_t, a_t)$$

All state transitions are deterministic, so only the actions generated by the LLM as text commands determine the agent's trajectory. Observations are likewise deterministic. We assume that the language model models an underlying policy, tapping into the capabilities under examination.

\subsection{Interaction}\label{subsec:interaction}
World state, representing a temporary subset of $S$, is stored as a set of fact tuples, describing both mutable states of entities and immutable states. Mutable states are \texttt{at}, \texttt{in}, \texttt{on}, \texttt{open} and \texttt{closed}. Immutable states describe entity types and categories and are used as conditions for actions (see Table~\ref{tab:actions}).

Actions are defined in the PDDL~\cite{fox2003pddl2} format, covering the state changes they enact (representing $T$). The action definition also contains a \citet{lark} grammar snippet that is used to form the parse-able input command grammar, feedback templates for success and individual failures (covering $O$) and a Clingo~\cite{GebserKKS2017clingo} encoding snippet of the state changes of the action for optimal solution solving (representing $T$ as well).

The player is not given a list of currently available actions but rather has to model the action space itself. Movement is only allowed to connected rooms (unlike in \citet{basavatia2024starlingselfsupervisedtrainingtextbased} and others, which allow ``teleportation'').

\textbf{Turn Limit}: The interaction is limited to 50 turns, and reaching the limit is recorded as aborting that episode.

\textbf{Formatting}: An output produced by LLM has to follow the format below (starting with the prompt symbol \textit{>}), e.g. \textit{> go hallway}:
\begin{verbatim}
    > "Natural language command"
\end{verbatim}

The episodes where generated outputs that do not follow this formatting are aborted.

\textbf{Parsing \& State Change}: The interpreter attempts to parse the command, and if it passes, the corresponding action is performed in the resolution phase. State change conditions are checked, and any resolution failure stops the process. If state change conditions are fulfilled, the game world state is updated accordingly, removing and adding facts in the world state. All changes in the world state are recorded. Lastly, the interpreter checks if changed states are part of the goal state set $G$ and returns achieved goal states $G_a$, textual feedback $o\in O$ and any failure that might have occurred in processing the input command to be recorded.

The player ends an episode by generating the ``done'' command. Once the episode ends, goal states achieved as of the last turn are recorded—meaning that goal states achieved intermittently are not considered for metrics.

\textbf{Exploration Tracking}: All commands, state changes, observed locations and objects are recorded for each turn, including errors while executing the commands. We label each action (inspired by \citet{KIRSH1994513}) for being \textit{epistemic} and \textit{pragmatic}. An action is \textit{epistemic} when it improves a player's perception of the game situation without directly progressing towards the goal\footnote{Kirsh and Maglio's example is moving a Tetris tile to the leftmost position to be sure about its position instead of moving it towards the location that is most beneficial to put it down in.} and \textit{pragmatic} when the action directly works towards reaching the goal. Then, we calculate \textit{epistemic gain} by counting how many world state facts the player newly observes as a result.

\section{Experimental Setup}

\subsection{Game Instances}\label{subsec:game-instances}

We create different set of instances to test incremental learning under uncontrolled (Basic, Synthetic words) and controlled (Pre-exploration, Inventory limit) conditions.

\subsubsection{Basic}
The core task in our experiments is a \textit{delivery task} where the agent is expected to deliver three objects to target receptacles in a typical home environment.

\noindent\textbf{Objects and rooms}: The basic instances contain 22 everyday household objects (e.g. plant, book), six common room types (e.g. kitchen, hallway) and seven staple IF actions (e.g. go, take). The complete lists are in Appendix~\ref{sec:basic-domain}. 

\noindent\textbf{Difficulty levels}: easy \& hard. The \textit{easy level} means that goal objects are not inside closed containers (easily accessible, immediately observable once the agent enters the room) and are located near the target receptacle in the initial world state. The \textit{hard level} means that goal objects are initially hidden in closed containers (e.g. cupboards, closets), where each needs to be delivered to a different target. There are also longer paths between initial goal object locations and their targets. Our hypothesis is that it is challenging for models to navigate multi-step actions to reach a goal because hidden objects require epistemic actions, and far away objects require keeping the objective in mind over several steps (increases the number of steps to reach goals).

\subsubsection{Inventory Limit}
Using the same settings described above, we create another configuration by setting a limit on how many objects can be carried by the agent at a time. We set the limit as \textit{two} objects. We want to analyse whether models lean towards strategic use of the resource ``inventory'' and are efficient when they have the limit. The set of objects and rooms in instances used here are the same as in \textit{Basic} experiment.

\subsubsection{Pre-Exploration}
We aim to test the effect of additional context at the start of the interaction. We provide the information about the rooms and which objects are there located there to the agent. This effectively inserts the layout of rooms needed for correct navigation and locations of task objects (or the containers holding them in 'hard' instances) into context. As this automated interaction is formatted like any correct interaction input and feedback, the pre-exploration sequence can be considered as a \textit{few-shot in-context learning}. Pre-exploration sequences are six to eight ``go'' actions. We want to analyse whether models get more efficient (compared to the \textit{Basic} experiment) in completing the goals. The set of objects and rooms in instances used here are the same as in \textit{Basic} experiment.

\subsubsection{Synthetic Words}
We create another experiment set by replacing words in the home delivery task with pseudo-English words using scripts by \citet{eisenschlos2023winodict} while retaining an unaltered basic set of common IF actions. The goal here is to assess whether agents can solve the task incrementally by relying solely on the feedback from the environment. These words are unknown to any LLM. Thus, embedded common-sense knowledge does little affect performance and focuses more on generalisation abilities. We defined three difficulty levels: easy, medium \& hard.

\textbf{Easy level}: Three objects and one action word are replaced with pseudo-English words. A short explanation of the replaced action word is provided. For example, we apply the following replacements: \textit{bedroom} $\rightarrow$ \textit{enticed}, \textit{book} $\rightarrow$ \textit{decte}, \textit{shelf} $\rightarrow$ \textit{stord}. The verb \textit{put} is replaced with \textit{aphon}, and the explanation ``In addition to common actions, you can aphon. To aphon is to physically place something somewhere.'' is provided in the initial prompt. The correct command to \textit{put the book on the shelf} is then \textit{> aphon decte (on) stord}.

\textbf{Medium level}: Nine objects and three action words are replaced with pseudo-English words. The replaced synthetic actions words are listed in the initial prompt without explanations, such as ``In addition to common actions, you can inate, pante and eness''. In this case, \textit{eness} replaces \textit{close}, \textit{pante} replaces \textit{put} and \textit{inate} replaces \textit{open}.

\textbf{Hard level}: This variant uses pseudo-English words for all entity and room nouns, state adjectives and action verbs, with the task being to change the states of entities using the verbs, removing the possibility of relying on common-sense knowledge. As we expected, the ubiquity of synthetic words would provide great difficulty in itself, but these instances contain only four rooms in a simple linear arrangement. Object state interaction complexity ranges from single state change (\textit{> mator subst} directly resulting in \textit{the subst is now dent}, fulfilling a goal state), via binary state sets (\textit{> unbal diale} changes the \textit{unsust-able}' \textit{diale} object from being \textit{unsust} to being \textit{exper}), to state sets with three states, requiring the use of two actions in the correct order to bring an object into a target state.

Our intention here is to validate models' reliance on common-sense knowledge seen during their training phases. These experiments require the models to apply in-context learning to uncover the meanings of synthetic words by interacting with the environment and getting feedback on their actions.

\subsubsection{Number of Instances}
In total, we have nine experiments for each variant described above: Basic-easy, Basic-hard, Basic-easy-limit-two, Basic-hard-limit-two, Basic-easy-preexplore, Basic-hard-preexplore, Synthetic-words-easy, Synthetic-words-medium and Synthetic-words-hard.
Each experiment has \textit{16} instances, corresponding to \textit{144} instances in total.

\subsection{Metrics}
\label{sec:metrics}

\textbf{Framework-specific metrics}: The clembench framework includes two main metrics: \textit{Played} \& \textit{Quality Score}. The game finishes successfully only when a model produces \textit{> done} as the last action and all goals have been achieved. The game is \textit{aborted} when a model does not follow formatting requirements (see Section~\ref{subsec:interaction}) or reaches the \textit{maximum turn limit}, which is 50. Played is the ratio of instances that were not aborted. Quality Score measures how many episodes have all their goal states reached at the end. Producing the \textit{> done} action command without achieving all goal states is considered a \textit{lost} episode. In cases where all goal states are achieved and \textit{> done} is also generated, then the episode is \textit{successful}.

Finally, to rank the benchmarked models, the framework includes a metric called \textit{clemscore}, the macro-average \textit{Quality Score} multiplied by the macro-average proportion of \textit{Played} games across all experiments.

\textbf{Game-specific metrics}:
We have a specific metric to keep track of achieved goals. It is the ratio between achieved goal states $G_a$ and all goal states $G$: Goal Success Rate (GSR) $ = \frac{|G_a|}{|G|}\times 100$.

\subsection{Evaluated Models}
We evaluated open-weight and commercial models (with \textit{temp=0}). We included recent commercial models such as: \textit{o3-mini} (Jan~'25), \textit{GPT-4o} (Aug~'24) \textit{Claude-3-7} (Sonnet, Feb~'25). We also included recent open-weight models: \textit{Llama-3.1} (8B, 70B)~\citep{llama31}, \textit{Llama-3.3} (70B), \textit{Qwen2.5} (Coder-32B, 72B)~\citep{qwen25}, and \textit{Deepseek-v3}~\citep{deepseekv3}. We used the APIs of the respective commercial models. We ran open-weight models on two NVIDIA A100 GPUs. Deepseek-v3 was run via the OpenRouter API.

\begin{table}[ht!]
\footnotesize
    \centering

\begin{tabular}{lcccc}
\hline
\textbf{Model} & \makecell{\textbf{clem} \\\textbf{score}} & \makecell{\textbf{Quality} \\\textbf{Score}} & \textbf{\% Played} & \makecell{\textbf{Goal} \\\textbf{Rate}} \\
\hline
Claude-3.7 & \textbf{86.4} & \textbf{88.2} & \textbf{97.9} & \textbf{90.5} \\
o3-mini & 68.7 & 79.2 & 86.8 & 85.9 \\
GPT-4o & 52.7 & 56.2 & 93.8 & 75.9 \\
Llama-3.1-70B & 43.8 & 49.2 & 89.2 & 66.8 \\
Llama-3.3-70B & 40.1 & 43.3 & 92.5 & 64.4 \\
Deepseek-v3 & 39.8 & 46.5 & 85.4 & 64.8 \\
Qwen2-72B & 15.9 & 28.8 & 55.4 & 47.5 \\
Qwen2.5-32B & 12.0 & 25.0 & 47.9 & 47.8 \\
Qwen2.5-72B & 11.2 & 21.2 & 52.9 & 44.9 \\
Llama-3.1-8B & 8.2 & 20.6 & 39.9 & 35.7 \\
\hline
\end{tabular}
    \caption{Overall benchmark scores for models.}
    \label{tab:clemscores}
\end{table}

\section{Results}
\subsection{Overall Comparison}
Table \ref{tab:clemscores} shows the main scores for the benchmarked models. Larger models achieve higher quality scores and better conform to the prompted output format (yielding higher \% Played). Most commercial models achieve higher scores than open-weight models (8.9 points between \textit{GPT-4o} and \textit{Llama-3.1-70B}), with \textit{Claude-3.7} scoring higher than the next best model by at least 9 points on the three major metrics. Another observation is that all high-ranking models can play the game (follow the instructions) but lack performance in solving the task. Next, we analyse the cases deeper to uncover which factors contribute to low scores.

\begin{table*}[ht]
\centering
\footnotesize

\begin{tabular}{|ll|cc|cc|cc|cc|cc|cc|}
\hline
\multicolumn{2}{|c|}{\textbf{Experiment}} & \multicolumn{2}{c|}{Claude-3.7} & \multicolumn{2}{c|}{o3-mini} & \multicolumn{2}{c|}{GPT-4o} & \multicolumn{2}{c|}{LM-3.1} & \multicolumn{2}{c|}{LM-3.3} & \multicolumn{2}{c|}{DS-v3} \\
\cline{3-14}
\multicolumn{2}{|c|}{} & Q & P & Q & P & Q & P & Q & P & Q & P & Q & P \\ 
\hline
\multicolumn{1}{|l|}{\multirow{2}{*}{\textbf{Basic}}}
& \cellcolor{basicLight}easy & \cellcolor{basicLight}75.0 & \cellcolor{basicLight}100 & \cellcolor{basicLight}75.0 & \cellcolor{basicLight}87.5 & \cellcolor{basicLight}68.7 & \cellcolor{basicLight}100 & \cellcolor{basicLight}62.5 & \cellcolor{basicLight}100 & \cellcolor{basicLight}56.2 & \cellcolor{basicLight}100 & \cellcolor{basicLight}62.5 & \cellcolor{basicLight}100 \\ 
\multicolumn{1}{|l|}{}
& \cellcolor{basicDark}hard & \cellcolor{basicDark}81.2 & \cellcolor{basicDark}93.7 & \cellcolor{basicDark}87.5 & \cellcolor{basicDark}93.7 & \cellcolor{basicDark}43.7 & \cellcolor{basicDark}87.5 & \cellcolor{basicDark}31.2 & \cellcolor{basicDark}87.5 & \cellcolor{basicDark}18.7 & \cellcolor{basicDark}93.7 & \cellcolor{basicDark}56.2 & \cellcolor{basicDark}87.5 \\ 
\hline
\multicolumn{1}{|l|}{\multirow{2}{*}{\textbf{Pre-Exploration}}}
& \cellcolor{preExploreLight}easy & \cellcolor{preExploreLight}75.0 & \cellcolor{preExploreLight}100 & \cellcolor{preExploreLight}75.0 & \cellcolor{preExploreLight}93.7 & \cellcolor{preExploreLight}87.5 & \cellcolor{preExploreLight}100 & \cellcolor{preExploreLight}68.7 & \cellcolor{preExploreLight}100 & \cellcolor{preExploreLight}68.7 & \cellcolor{preExploreLight}100 & \cellcolor{preExploreLight}93.7 & \cellcolor{preExploreLight}100 \\ 
\multicolumn{1}{|l|}{}
& \cellcolor{preExploreDark}hard & \cellcolor{preExploreDark}93.7 & \cellcolor{preExploreDark}93.7 & \cellcolor{preExploreDark}75.0 & \cellcolor{preExploreDark}75.0 & \cellcolor{preExploreDark}37.5 & \cellcolor{preExploreDark}87.5 & \cellcolor{preExploreDark}50.0 & \cellcolor{preExploreDark}93.7 & \cellcolor{preExploreDark}31.2 & \cellcolor{preExploreDark}100 & \cellcolor{preExploreDark}37.5 & \cellcolor{preExploreDark}56.2 \\ 
\hline
\multicolumn{1}{|l|}{\multirow{2}{*}{\textbf{Inventory Limit}}}
& \cellcolor{inventoryLight}easy & \cellcolor{inventoryLight}81.2 & \cellcolor{inventoryLight}100 & \cellcolor{inventoryLight}87.5 & \cellcolor{inventoryLight}100 & \cellcolor{inventoryLight}68.7 & \cellcolor{inventoryLight}100 & \cellcolor{inventoryLight}62.5 & \cellcolor{inventoryLight}100 & \cellcolor{inventoryLight}50.0 & \cellcolor{inventoryLight}100 & \cellcolor{inventoryLight}68.7 & \cellcolor{inventoryLight}100 \\ 
\multicolumn{1}{|l|}{}
& \cellcolor{inventoryDark}hard & \cellcolor{inventoryDark}93.7 & \cellcolor{inventoryDark}93.7 & \cellcolor{inventoryDark}93.7 & \cellcolor{inventoryDark}93.7 & \cellcolor{inventoryDark}31.2 & \cellcolor{inventoryDark}87.5 & \cellcolor{inventoryDark}37.5 & \cellcolor{inventoryDark}87.5 & \cellcolor{inventoryDark}31.2 & \cellcolor{inventoryDark}93.7 & \cellcolor{inventoryDark}31.2 & \cellcolor{inventoryDark}56.2 \\ 
\hline
\multicolumn{1}{|l|}{\multirow{3}{*}{\textbf{Synthetic Words}}}
& \cellcolor{syntheticLight}easy & \cellcolor{syntheticLight}100 & \cellcolor{syntheticLight}100 & \cellcolor{syntheticLight}87.5 & \cellcolor{syntheticLight}100 & \cellcolor{syntheticLight}81.2 & \cellcolor{syntheticLight}100 & \cellcolor{syntheticLight}68.7 & \cellcolor{syntheticLight}93.7 & \cellcolor{syntheticLight}43.7 & \cellcolor{syntheticLight}93.7 & \cellcolor{syntheticLight}43.7 & \cellcolor{syntheticLight}100 \\ 
\multicolumn{1}{|l|}{}
& \cellcolor{syntheticMedium}medium & \cellcolor{syntheticMedium}100 & \cellcolor{syntheticMedium}100 & \cellcolor{syntheticMedium}43.7 & \cellcolor{syntheticMedium}43.7 & \cellcolor{syntheticMedium}43.7 & \cellcolor{syntheticMedium}81.2 & \cellcolor{syntheticMedium}37.5 & \cellcolor{syntheticMedium}50.0 & \cellcolor{syntheticMedium}18.7 & \cellcolor{syntheticMedium}50.0 & \cellcolor{syntheticMedium}25.0 & \cellcolor{syntheticMedium}81.2 \\ 
\multicolumn{1}{|l|}{}
& \cellcolor{syntheticDark}hard & \cellcolor{syntheticDark}93.7 & \cellcolor{syntheticDark}100 & \cellcolor{syntheticDark}87.5 & \cellcolor{syntheticDark}93.7 & \cellcolor{syntheticDark}43.7 & \cellcolor{syntheticDark}100 & \cellcolor{syntheticDark}12.5 & \cellcolor{syntheticDark}87.5 & \cellcolor{syntheticDark}43.75 & \cellcolor{syntheticDark}93.75 & \cellcolor{syntheticDark}0.0 & \cellcolor{syntheticDark}87.5 \\ 
\hline
\end{tabular}

\caption{Detailed results across different experiments. Only high performing six LLMs were selected. The values are \textit{Quality Score} (Q) and \textit{\% Played} (P) separated by the \textit{/} (slash sign) for each experiment. \textit{LM-3.1} $\rightarrow$ Llama-3.1-70B, \textit{LM-3.3} $\rightarrow$ Llama-3.3-70B, DS-v3 $\rightarrow$ Deepseek-v3 }
\label{table:detailed-experiments}
\end{table*}

\subsection{In-depth Analysis}

Table~\ref{table:detailed-experiments} presents the \textit{Quality} and \textit{\% Played} values for each experiment. We selected the top-performing six models. The results for other models can be found in Table~\ref{tab:experiment-results-low}. Next, we break down the results across each experiment.

\textbf{Basic Navigation \& Task Solving}: In the \textit{easy} setting, most models seem to get somewhat adequate performance (>=50). Notably, among the compared top six models, \textit{o3-mini} is the only model without a full \% Played score in this experiment, leading to it not matching the performance of the best model, \textit{Claude-3.7}. However, \textit{o3-mini} achieves the highest scores in the \textit{hard} setting, with it and especially \textit{Claude-3.7} being impacted far less by the increased task complexity.

\begin{figure}[ht]
    \centering
    \includegraphics[width=1.0\linewidth]{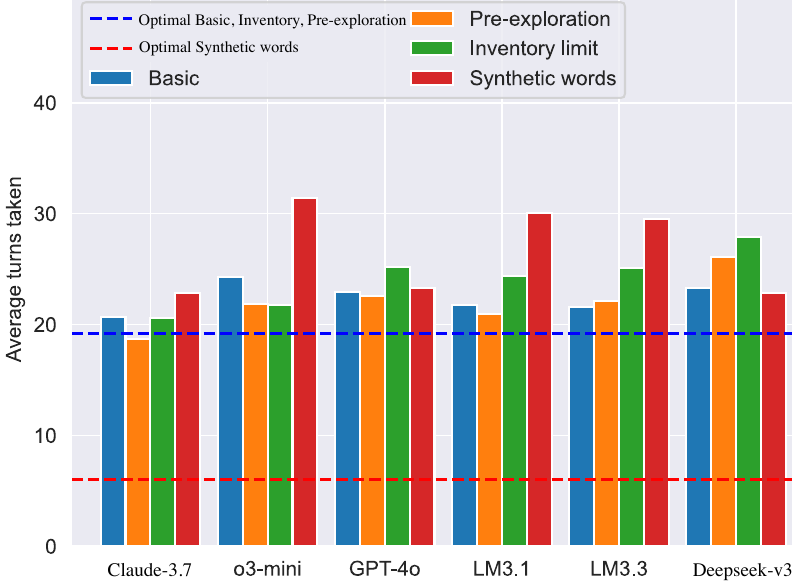}
    \caption{Average number of turns taken by each model for each experiment category and average optimal number of turns for each category are also included (dotted lines).}
    \label{fig:speed-selected-models}
\end{figure}

\textbf{Effects of Pre-Exploration}: This experiment aims to see whether models increase their performance since initial rooms and certain objects and containers are revealed. When comparing the results in the \textit{Basic} experiment. The unexpected behaviour is with \textit{o3-mini} and \textit{GPT-4o} on the \textit{hard}, where their performance worsens despite the provided pre-exploration context. Figure~\ref{fig:speed-selected-models} visualises the average number of turns it takes for each experiment category. The figure also provides the optimal number of turns (averaged). We can clearly observe that pre-exploration indeed helps the high-scoring models (Claude-3.7) to be more efficient (requiring less number of turns) and yield better performance when compared with the \textit{Basic} experiments.

\textbf{Inventory Limit}: When compared with the \textit{Basic} experiments, the performance of \textit{Claude-3.7} and \textit{o3-mini} models get better while others yield similar performance. Figure~\ref{fig:speed-selected-models} visualises similar patterns where they do not need more turns than the \textit{Basic} experiments. \textit{o3-mini} even reduces the number of turns with the limit. It suggests that these models can navigate the inventory limit while retaining or even increasing performance. However, lower-performing models (GPT-4o, LLama models, Deepseek-v3) tend to take much longer than the \textit{Basic} experiment. It shows that these models get affected by the inventory limit not much on the general task solving but having lower efficiency.

\textbf{Synthetic Words}: In the \textit{easy} setting, the smaller models (\textit{Llama-3.1-8B}, \textit{Qwen2.5-32B-Coder}) yield the worst performance showing that the required incremental learning abilities are severely lacking in smaller models. All commercial models \textit{o3-mini}, \textit{GPT-4o}, \textit{Claude-3.7}, \textit{GPT-4o} show an improvement over the \textit{Basic} experiment results. \textit{Claude-3.7} even achieves full scores, making it the best model in this experiment.

The larger amount of synthetic words in the \textit{medium} setting does lead to a drastic drop (expected by the design of the experiment) in scores compared to the \textit{easy} experiment, lowering scores by at least half. However, \textit{Claude-3.7} still yields the perfect scores, again.

Performance impact diverges between models for the \textit{hard} setting, where we expected the task to be more challenging for the models than the \textit{medium} setting. While \textit{Llama-3.1-70B} and \textit{DeepSeek-v3} (0.0 Quality Score) do worse when confronted with many synthetic words, \textit{o3-mini} performs almost as well as in the \textit{easy} experiment and \textit{Claude-3.7} achieves high performance again (fails only in a single episode). 

Figure~\ref{fig:speed-selected-models} shows that none of the models are that efficient (requiring more turns than the optimal solution) as they were with other experiments. It suggests that, as expected, the models need more turns to infer the meaning of words. Claude-3.7 is not only the best performing model but also the most efficient one in these experiments.

\begin{figure*}[ht!]
    \begin{minipage}{0.50\textwidth}
    \includegraphics[width=\linewidth]{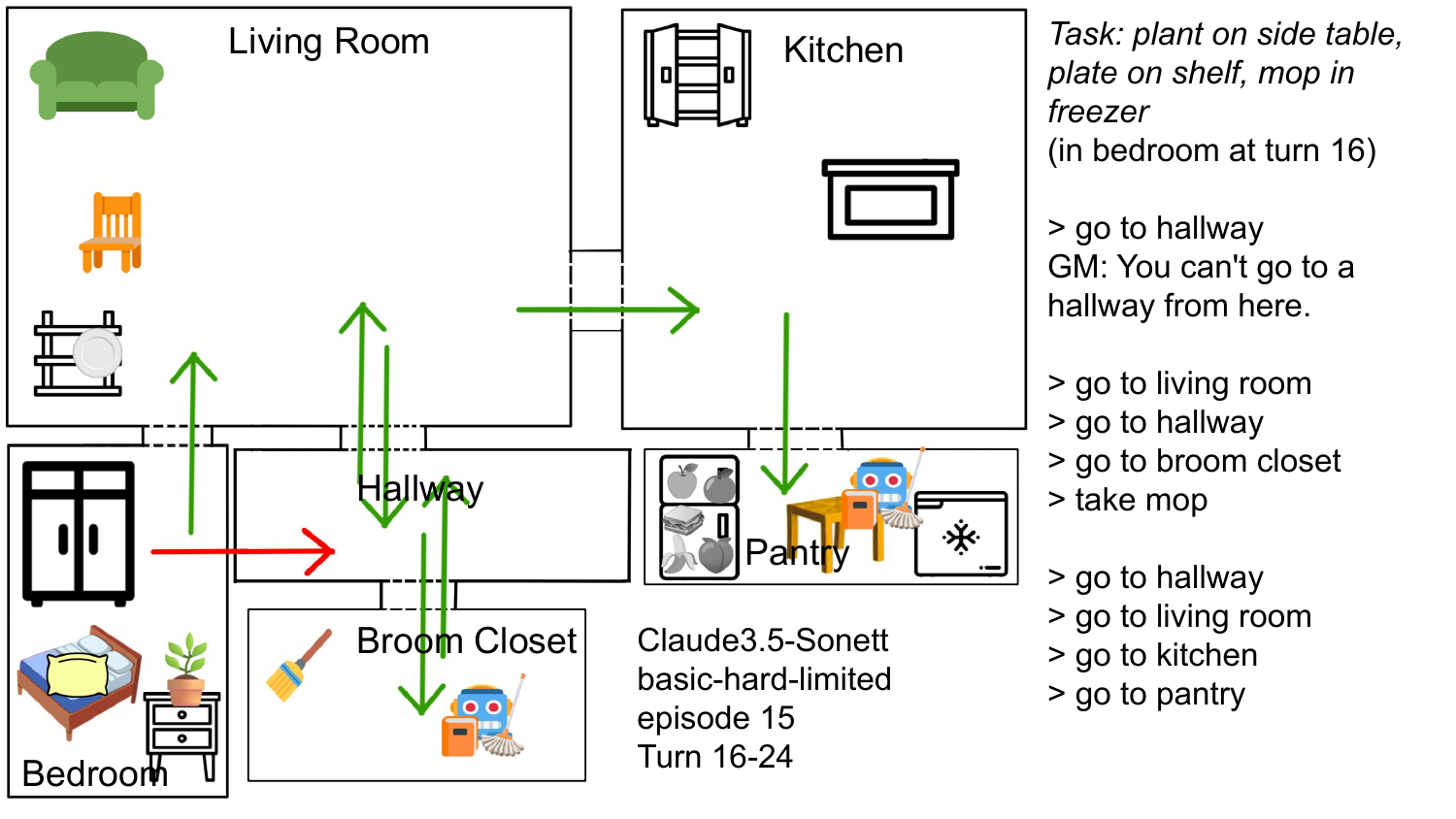}
    \caption{\textit{Claude-3.7} correcting navigation.}
    \label{fig:claude-navigation}
    \includegraphics[width=\linewidth]{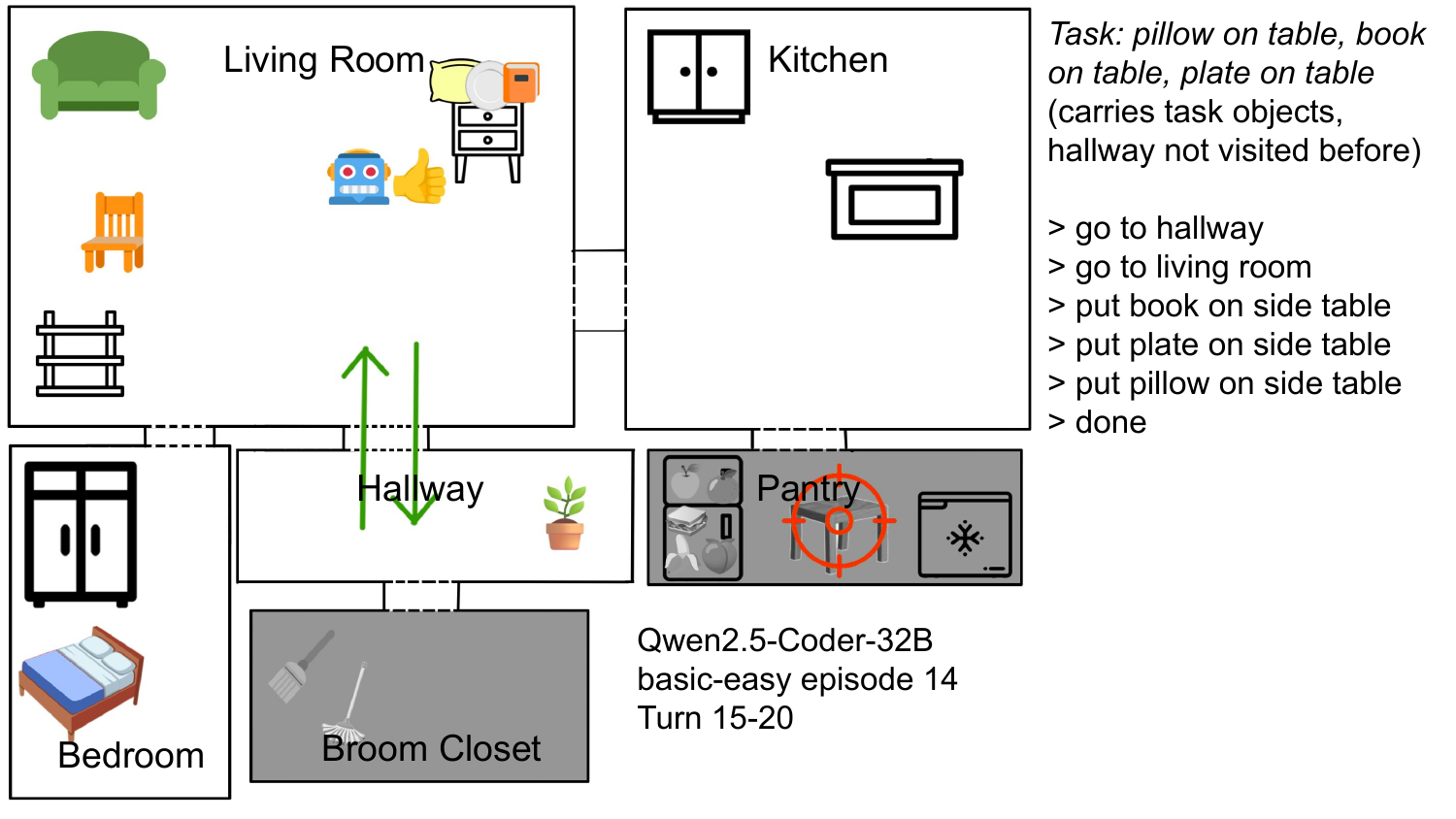}
    \caption{\textit{Qwen2.5-32}B exploring insufficiently.}
    \label{fig:qwen32-exploration}
    \end{minipage}%
    \begin{minipage}{0.50\textwidth}
    \includegraphics[width=\linewidth]{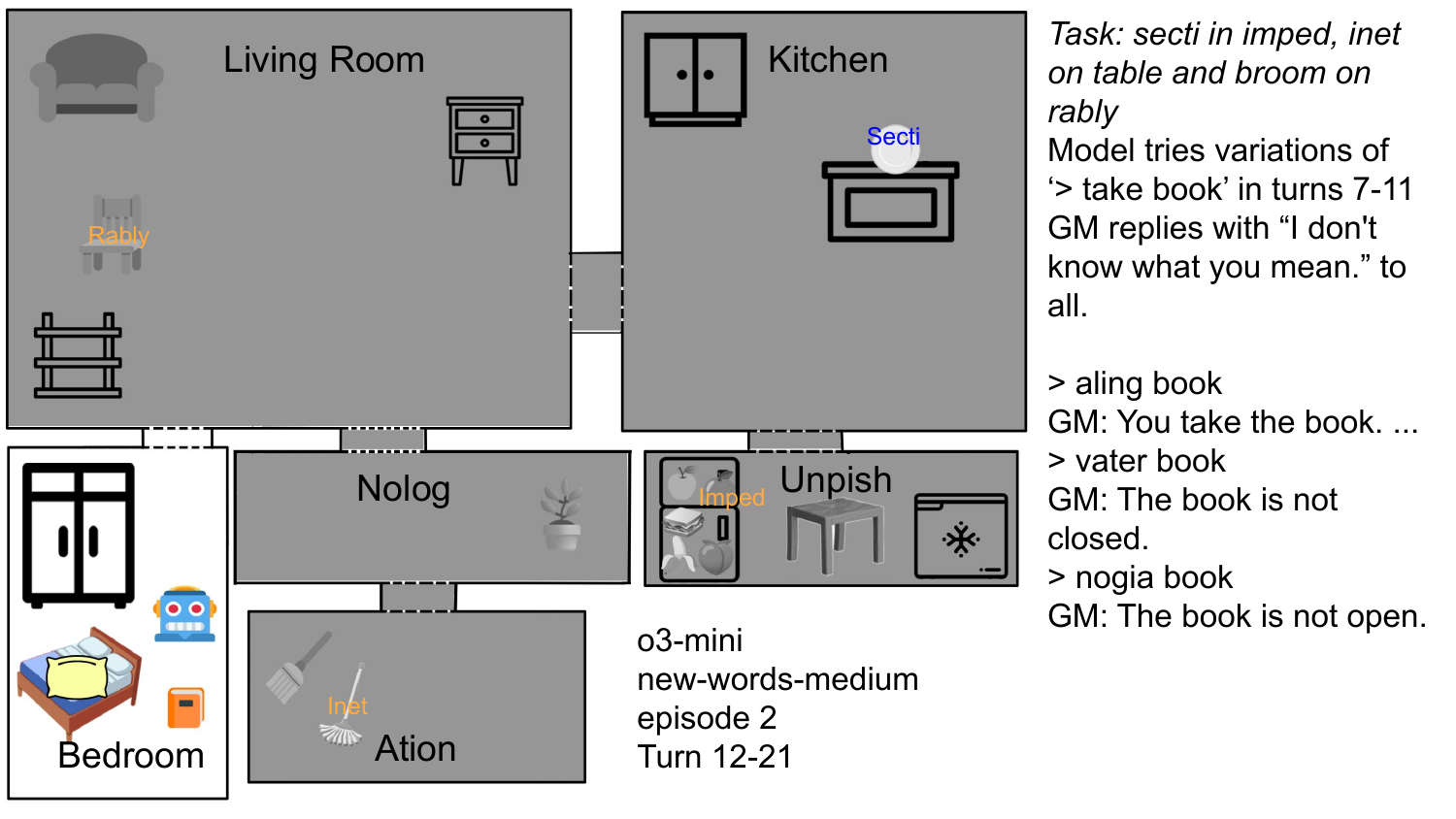}
    \caption{\textit{o3-mini} exploring synthetic actions.}
    \label{fig:o3-mini-new-words1}
    \includegraphics[width=\linewidth]{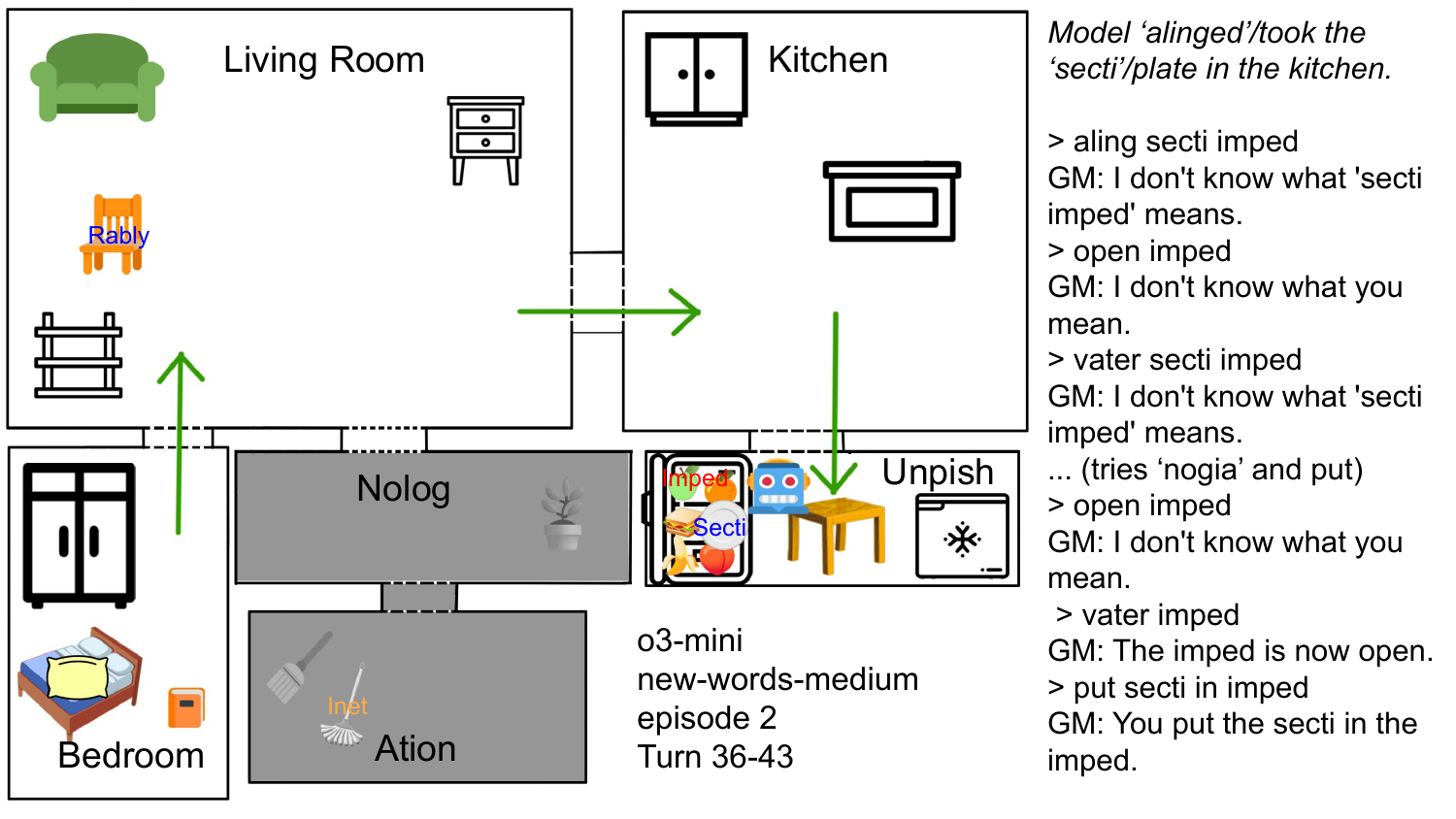}
    \caption{\textit{o3-mini} misapplying synthetic actions.}
    \label{fig:o3-mini-new-words2}
\end{minipage}
\end{figure*}

\subsection{Qualitative Analysis}
\textbf{Navigation and self-correction}: 
All models make navigation errors despite the observation feedback mentioning passages to all connected rooms every time the player enters a room. While lower-performing models repeat this type of failure more in individual episodes, higher-performing models acquire the game's rule that allows movement only to connected rooms and self-correct their navigation after receiving feedback. This type of failure most often occurs in the turn after a task object has been picked up or delivered. Figure \ref{fig:claude-navigation} illustrates how \textit{Claude-3.7} attempts to "> go to hallway" from the unconnected bedroom (red arrow) and is told that this is not possible. It then respects the connection requirement for the rest of the episode and goes to the \textit{broom closet} to \textit{take the mop} and further \textit{to the pantry with the freezer} without attempting to go to these known rooms directly.

\textbf{Insufficient exploration}: Figure \ref{fig:qwen32-exploration} illustrates how \textit{Qwen2.5-Coder-32B} regularly checks connected rooms, e.g. \textit{hallway}, which largely contributes to its performance. However, it does not do this thoroughly enough, missing the \textit{table in the pantry} and incorrectly placing the task objects on the side \textit{table}, as most models do.

\textbf{Synthetic words \& in-context learning}: Figure \ref{fig:o3-mini-new-words1} shows a typical behaviour of well-performing models in the synthetic words experiment: After attempts to take the book (regardless of it not being needed for the task), \textit{o3-mini} tries the synthetic word actions provided in the initial prompt. The environment feedback contains the information necessary to learn that ``aling'' means to ``take'', ``vater'' means to ``open'' and ``nogia'' means to ``close''. Later in the same episode, shown in Figure~\ref{fig:o3-mini-new-words2}, \textit{o3-mini} fails to use the synthetic verbs properly. It eventually produces the right action ``vater imped'', allowing it to fulfil one of the three task goals, but the episode ends due to reaching the  turn limit.

\section{Discussion}

\textbf{Small models lack generalisation capabilities}: models that perform worst in the synthetic word experiments lack the ability to generalise beyond the content of their training data, with the introduction of even a few unseen token sequences in context disrupting their performance.

\textbf{Large models show individual differences for common-sense knowledge vs. in-context learning (ICL)}: the mixed performance between larger models indicates substantial differences in ICL and generalisation capabilities between individual models. While these models do not fail like the smaller ones, they are distracted by encountering synthetic words, eventually fail to solve the tasks.

\textbf{Automatic provision of information can be detrimental}: the negative impact of the navigation head-start in the pre-exploration experiments compared with the basic experiments is unexpected, as it should provide a large part of the necessary exploration and the situation state information needed to solve the task better. However, we assume that the observed performance is an effect of ICL as well: The pre-exploration sequence essentially serves as few-shot examples \textit{to not interact with task-relevant objects} when they are encountered, which is opposed to the behaviour the tested models show without it, interacting with task objects immediately and thus directly pursuing task goals.

\textbf{Situation complexity is more important than words}: the difference in scores in the \textit{easy} and \textit{medium} variants suggests that strong common sense knowledge impedes the ability to learn new information for all models except \textit{Claude-3.7}, which excels in this acquisition. The higher scores some models achieve in the \textit{hard} synthetic words experiment are due to the relatively low complexity of the situation, requiring less navigational exploration and lacking task objects that need to be revealed through interaction. Models that can learn the meaning of arbitrary synthetic words from the feedback provided when interacting with these words can thus solve these tasks better, as shown by the high scores of \textit{o3-mini} and \textit{Claude-3.7}.

\textbf{Human baseline}: we conducted a small study and played five instances from each experiment and measured 100 points in all metrics while Claude-3.7 scoring very close to it (see Table~\ref{tab:experiment-results-overview-human}).

\section{Conclusion}

Our experimental results indicate that performance increases with model size, progressing from generally bad situation modelling in smaller models to a middle ground of good situation modelling but frequent interaction failures, to only a few larger models fulfilling the given task in more than two thirds of cases. Incremental learning abilities of models were tested even further with the synthetic words experiments and showed that the gap between commercial and open-weight models is still quite big. Models relying more on their embedded common-sense knowledge performed worse and were less efficient that those models that are capable of incrementally figuring out the task and applying in-context learning.

\section*{Limitations}

The current study is restricted to only English in its current state. While we have yet to do this, translating the prompts and adapting the underlying grammar entries is possible for other languages, too.

The performance we measured here may not transfer to other modalities with more sophisticated demands, like visually or physically embodied agents or robots. \citet{alfworld} found that while training in text-only environments is faster and less resource-intensive than training in the AI2Thor framework, agents trained in text-only environments struggled to adapt to the requirements of more complex embodiment properly.

\section*{Ethics Statement}

In academic research, using paid proprietary APIs with underlying models about which little is known (training data, model architecture) is less than ideal. Currently, the models benchmarked here are the high-performing ones that are commercially used. We hope that more open models with high performance will be released soon and that proper research can be done on them.

Models that may be used in agents, possibly embodied as robots, being able to acquire replacement lexicons or entirely new semantics simply through ICL bears the possibility to circumvent safety measures from model creators and providers. We do not condone any such attempts based on our findings on the amenability of certain models to this kind of manipulation, commonly known as jailbreaks, specially when LLM agents are capable of interacting with computer systems or controlling robots.

\bibliography{custom}

\begin{thebibliography}{34}
\providecommand{\natexlab}[1]{#1}

\bibitem[{Basavatia et~al.(2024)Basavatia, Murugesan, and
  Ratnakar}]{basavatia2024starlingselfsupervisedtrainingtextbased}
Shreyas Basavatia, Keerthiram Murugesan, and Shivam Ratnakar. 2024.
\newblock \href {https://arxiv.org/abs/2406.05872} {Starling: Self-supervised
  training of text-based reinforcement learning agent with large language
  models}.
\newblock \emph{Preprint}, arXiv:2406.05872.

\bibitem[{Brown et~al.(2020)Brown, Mann, Ryder, Subbiah, Kaplan, Dhariwal,
  Neelakantan, Shyam, Sastry, and et~al.}]{DBLP:conf/nips/BrownMRSKDNSSAA20}
Tom~B. Brown, Benjamin Mann, Nick Ryder, Melanie Subbiah, Jared Kaplan,
  Prafulla Dhariwal, Arvind Neelakantan, Pranav Shyam, Girish Sastry, and
  et~al. 2020.
\newblock \href
  {https://proceedings.neurips.cc/paper/2020/hash/1457c0d6bfcb4967418bfb8ac142f64a-Abstract.html}
  {Language models are few-shot learners}.
\newblock In \emph{Advances in Neural Information Processing Systems 33: Annual
  Conference on Neural Information Processing Systems 2020, NeurIPS 2020,
  December 6-12, 2020, virtual}.

\bibitem[{Chalamalasetti et~al.(2023)Chalamalasetti, Götze, Hakimov,
  Madureira, Sadler, and Schlangen}]{clembench2023}
Kranti Chalamalasetti, Jana Götze, Sherzod Hakimov, Brielen Madureira, Philipp
  Sadler, and David Schlangen. 2023.
\newblock \href {https://arxiv.org/abs/2305.13455} {Clembench: Using game play
  to evaluate chat-optimized language models as conversational agents}.
\newblock \emph{Preprint}, arXiv:2305.13455.

\bibitem[{C{\^o}t{\'e} et~al.(2019)C{\^o}t{\'e}, K{\'a}d{\'a}r, Yuan, Kybartas,
  Barnes, Fine, Moore, Hausknecht, El~Asri, Adada, Tay, and
  Trischler}]{textworld}
Marc-Alexandre C{\^o}t{\'e}, {\'A}kos K{\'a}d{\'a}r, Xingdi Yuan, Ben Kybartas,
  Tavian Barnes, Emery Fine, James Moore, Matthew Hausknecht, Layla El~Asri,
  Mahmoud Adada, Wendy Tay, and Adam Trischler. 2019.
\newblock Textworld: A learning environment for text-based games.
\newblock In \emph{Computer Games}, pages 41--75, Cham. Springer International
  Publishing.

\bibitem[{Cui et~al.(2025)Cui, Yuan, Xiao, Ammanabrolu, and
  C\^ot\'e}]{cui2025tales}
Christopher Cui, Xingdi Yuan, Ziang Xiao, Prithviraj Ammanabrolu, and
  Marc-Alexandre C\^ot\'e. 2025.
\newblock \href {https://arxiv.org/abs/2504.14128} {Tales: Text-adventure
  learning environment suite}.
\newblock \emph{arXiv preprint arXiv:2504.14128}.

\bibitem[{DeepSeek-AI et~al.(2024)DeepSeek-AI, Liu, Feng, Xue, Wang, Wu, Lu,
  and et~al.}]{deepseekv3}
DeepSeek-AI, Aixin Liu, Bei Feng, Bing Xue, Bingxuan Wang, Bochao Wu, Chengda
  Lu, and et~al. 2024.
\newblock \href {https://arxiv.org/abs/2412.19437} {Deepseek-v3 technical
  report}.
\newblock \emph{Preprint}, arXiv:2412.19437.

\bibitem[{Eisenschlos et~al.(2023)Eisenschlos, Cole, Liu, and
  Cohen}]{eisenschlos2023winodict}
Julian~Martin Eisenschlos, Jeremy~R Cole, Fangyu Liu, and William Cohen. 2023.
\newblock Winodict: Probing language models for in-context word acquisition.
\newblock In \emph{Proceedings of the 17th Conference of the European Chapter
  of the Association for Computational Linguistics}, pages 94--102.

\bibitem[{Fox and Long(2003)}]{fox2003pddl2}
Maria Fox and Derek Long. 2003.
\newblock \href {https://www.jair.org/index.php/jair/article/view/10352}
  {Pddl2. 1: An extension to pddl for expressing temporal planning domains}.
\newblock \emph{Journal of artificial intelligence research}, 20:61--124.

\bibitem[{Gebser et~al.(2017)Gebser, Kaminski, Kaufmann, and
  Schaub}]{GebserKKS2017clingo}
Martin Gebser, Roland Kaminski, Benjamin Kaufmann, and Torsten Schaub. 2017.
\newblock \href
  {https://www.cambridge.org/core/journals/theory-and-practice-of-logic-programming/article/abs/multishot-asp-solving-with-clingo/FAED3429900D84CDD5155326A36548F2}
  {Multi-shot {ASP} solving with clingo}.
\newblock \emph{CoRR}, abs/1705.09811.

\bibitem[{Gioacchini et~al.(2024)Gioacchini, Siracusano, Sanvito, Gashteovski,
  Friede, Bifulco, and Lawrence}]{gioacchini2024agentquest}
Luca Gioacchini, Giuseppe Siracusano, Davide Sanvito, Kiril Gashteovski, David
  Friede, Roberto Bifulco, and Carolin Lawrence. 2024.
\newblock Agentquest: A modular benchmark framework to measure progress and
  improve llm agents.
\newblock \emph{arXiv preprint arXiv:2404.06411}.

\bibitem[{Gragera and Pozanco(2023)}]{gragera2023exploring}
Alba Gragera and Alberto Pozanco. 2023.
\newblock Exploring the limitations of using large language models to fix
  planning tasks.
\newblock In \emph{ICAPS Workshop on Knowledge Engineering for Planning and
  Scheduling (KEPS)}.

\bibitem[{Grattafiori et~al.(2024)Grattafiori, Dubey, Jauhri, Pandey, Kadian,
  and et~al.}]{llama31}
Aaron Grattafiori, Abhimanyu Dubey, Abhinav Jauhri, Abhinav Pandey, Abhishek
  Kadian, and et~al. 2024.
\newblock \href {https://arxiv.org/abs/2407.21783} {The llama 3 herd of
  models}.
\newblock \emph{Preprint}, arXiv:2407.21783.

\bibitem[{Hausknecht et~al.(2020)Hausknecht, Ammanabrolu, C{\^o}t{\'e}, and
  Yuan}]{hausknecht2020interactive}
Matthew Hausknecht, Prithviraj Ammanabrolu, Marc-Alexandre C{\^o}t{\'e}, and
  Xingdi Yuan. 2020.
\newblock Interactive fiction games: A colossal adventure.
\newblock In \emph{Proceedings of the AAAI Conference on Artificial
  Intelligence}, volume~34, pages 7903--7910.

\bibitem[{Ichter et~al.(2022)Ichter, Brohan, Chebotar, Finn, Hausman, Herzog,
  and et~al.}]{DBLP:conf/corl/IchterBCFHHHIIJ22}
Brian Ichter, Anthony Brohan, Yevgen Chebotar, Chelsea Finn, Karol Hausman,
  Alexander Herzog, and et~al. 2022.
\newblock \href {https://proceedings.mlr.press/v205/ichter23a.html} {Do as {I}
  can, not as {I} say: Grounding language in robotic affordances}.
\newblock In \emph{Conference on Robot Learning, CoRL 2022, 14-18 December
  2022, Auckland, New Zealand}, volume 205 of \emph{Proceedings of Machine
  Learning Research}, pages 287--318. {PMLR}.

\bibitem[{Jansen(2021)}]{jansen2021systematic}
Peter~A Jansen. 2021.
\newblock \href {https://arxiv.org/abs/2107.04132} {A systematic survey of text
  worlds as embodied natural language environments}.
\newblock \emph{arXiv preprint arXiv:2107.04132}.

\bibitem[{Kaelbling et~al.(1998)Kaelbling, Littman, and
  Cassandra}]{KAELBLING199899}
Leslie~Pack Kaelbling, Michael~L. Littman, and Anthony~R. Cassandra. 1998.
\newblock \href {https://doi.org/10.1016/S0004-3702(98)00023-X} {Planning and
  acting in partially observable stochastic domains}.
\newblock \emph{Artificial Intelligence}, 101(1):99--134.

\bibitem[{Kirsh and Maglio(1994)}]{KIRSH1994513}
David Kirsh and Paul Maglio. 1994.
\newblock \href {https://doi.org/10.1016/0364-0213(94)90007-8} {On
  distinguishing epistemic from pragmatic action}.
\newblock \emph{Cognitive Science}, 18(4):513--549.

\bibitem[{Kojima et~al.(2022)Kojima, Gu, Reid, Matsuo, and
  Iwasawa}]{DBLP:conf/nips/KojimaGRMI22}
Takeshi Kojima, Shixiang~Shane Gu, Machel Reid, Yutaka Matsuo, and Yusuke
  Iwasawa. 2022.
\newblock \href
  {http://papers.nips.cc/paper\_files/paper/2022/hash/8bb0d291acd4acf06ef112099c16f326-Abstract-Conference.html}
  {Large language models are zero-shot reasoners}.
\newblock In \emph{Advances in Neural Information Processing Systems 35: Annual
  Conference on Neural Information Processing Systems 2022, NeurIPS 2022, New
  Orleans, LA, USA, November 28 - December 9, 2022}.

\bibitem[{Kolve et~al.(2022)Kolve, Mottaghi, Han, VanderBilt, Weihs, Herrasti,
  Deitke, Ehsani, Gordon, Zhu, Kembhavi, Gupta, and
  Farhadi}]{kolve2022ai2thorinteractive3denvironment}
Eric Kolve, Roozbeh Mottaghi, Winson Han, Eli VanderBilt, Luca Weihs, Alvaro
  Herrasti, Matt Deitke, Kiana Ehsani, Daniel Gordon, Yuke Zhu, Aniruddha
  Kembhavi, Abhinav Gupta, and Ali Farhadi. 2022.
\newblock \href {https://arxiv.org/abs/1712.05474} {Ai2-thor: An interactive 3d
  environment for visual ai}.
\newblock \emph{Preprint}, arXiv:1712.05474.

\bibitem[{Lark parser()}]{lark}
Lark parser. 2024.
\newblock \href {https://github.com/lark-parser/lark} {[link]}.

\bibitem[{Ma et~al.(2024)Ma, Zhang, Zhu, Yang, Yang, Jin, Lan, Kong, and
  He}]{ma2024agentboard}
Chang Ma, Junlei Zhang, Zhihao Zhu, Cheng Yang, Yujiu Yang, Yaohui Jin,
  Zhenzhong Lan, Lingpeng Kong, and Junxian He. 2024.
\newblock Agentboard: An analytical evaluation board of multi-turn llm agents.
\newblock \emph{arXiv preprint arXiv:2401.13178}.

\bibitem[{Paglieri et~al.(2024)Paglieri, Cupia{\l}, Coward, Piterbarg,
  Wo{\l}czyk, Khan, Pignatelli, Kuci{\'n}ski, Pinto, Fergus, Foerster,
  Parker-Holder, and Rockt{\"a}schel}]{paglieri2024balrog}
Davide Paglieri, Bart{\l}omiej Cupia{\l}, Sam Coward, Ulyana Piterbarg, Maciej
  Wo{\l}czyk, Akbir Khan, Eduardo Pignatelli, {\L}ukasz Kuci{\'n}ski, Lerrel
  Pinto, Rob Fergus, Jakob~Nicolaus Foerster, Jack Parker-Holder, and Tim
  Rockt{\"a}schel. 2024.
\newblock Balrog: Benchmarking agentic llm and vlm reasoning on games.
\newblock \emph{arXiv preprint arXiv:2411.13543}.

\bibitem[{Qwen et~al.(2025)Qwen, Yang, Yang, Zhang, Hui, and et~al.}]{qwen25}
Qwen, An~Yang, Baosong Yang, Beichen Zhang, Binyuan Hui, and et~al. 2025.
\newblock \href {https://arxiv.org/abs/2412.15115} {Qwen2.5 technical report}.
\newblock \emph{Preprint}, arXiv:2412.15115.

\bibitem[{Schlangen(2023)}]{Schlangen-2023}
David Schlangen. 2023.
\newblock \href {https://doi.org/10.48550/arXiv.2302.08590} {What {A} situated
  language-using agent must be able to do: {A} top-down analysis}.
\newblock \emph{CoRR}, abs/2302.08590.

\bibitem[{Shi et~al.(2024)Shi, Wei, Xu, and Liang}]{shi2024why}
Zhenmei Shi, Junyi Wei, Zhuoyan Xu, and Yingyu Liang. 2024.
\newblock \href {https://openreview.net/forum?id=WOa96EG26M} {Why larger
  language models do in-context learning differently?}
\newblock In \emph{Forty-first International Conference on Machine Learning}.

\bibitem[{Shinn et~al.(2023)Shinn, Cassano, Gopinath, Narasimhan, and
  Yao}]{reflexion}
Noah Shinn, Federico Cassano, Ashwin Gopinath, Karthik Narasimhan, and Shunyu
  Yao. 2023.
\newblock \href
  {https://proceedings.neurips.cc/paper_files/paper/2023/file/1b44b878bb782e6954cd888628510e90-Paper-Conference.pdf}
  {Reflexion: language agents with verbal reinforcement learning}.
\newblock In \emph{Advances in Neural Information Processing Systems},
  volume~36, pages 8634--8652. Curran Associates, Inc.

\bibitem[{Shridhar et~al.(2021)Shridhar, Yuan, Côté, Bisk, Trischler, and
  Hausknecht}]{alfworld}
Mohit Shridhar, Xingdi Yuan, Marc-Alexandre Côté, Yonatan Bisk, Adam
  Trischler, and Matthew Hausknecht. 2021.
\newblock \href {https://arxiv.org/abs/2010.03768} {Alfworld: Aligning text and
  embodied environments for interactive learning}.
\newblock \emph{Preprint}, arXiv:2010.03768.

\bibitem[{Tan et~al.(2023)Tan, Kazemi, and Mihalcea}]{tan2023text}
Qinyue Tan, Ashkan Kazemi, and Rada Mihalcea. 2023.
\newblock \href {https://openreview.net/forum?id=2g4m5S_knF} {Text-based games
  as a challenging benchmark for large language models}.

\bibitem[{Tsai et~al.(2023)Tsai, Zhou, Liu, Li, Yu, and Mei}]{tsai2023can}
Chen~Feng Tsai, Xiaochen Zhou, Sierra~S Liu, Jing Li, Mo~Yu, and Hongyuan Mei.
  2023.
\newblock \href {https://arxiv.org/abs/2304.02868} {Can large language models
  play text games well? current state-of-the-art and open questions}.
\newblock \emph{arXiv preprint arXiv:2304.02868}.

\bibitem[{Valmeekam et~al.(2023)Valmeekam, Marquez, Olmo, Sreedharan, and
  Kambhampati}]{plan-bench}
Karthik Valmeekam, Matthew Marquez, Alberto Olmo, Sarath Sreedharan, and
  Subbarao Kambhampati. 2023.
\newblock \href
  {https://proceedings.neurips.cc/paper_files/paper/2023/file/7a92bcdede88c7afd108072faf5485c8-Paper-Datasets_and_Benchmarks.pdf}
  {Planbench: An extensible benchmark for evaluating large language models on
  planning and reasoning about change}.
\newblock In \emph{Advances in Neural Information Processing Systems},
  volume~36, pages 38975--38987. Curran Associates, Inc.

\bibitem[{Valmeekam et~al.(2024)Valmeekam, Stechly, and
  Kambhampati}]{valmeekam2024llms}
Karthik Valmeekam, Kaya Stechly, and Subbarao Kambhampati. 2024.
\newblock \href {https://openreview.net/forum?id=Gcr1Lx4Koz} {{LLM}s still
  can't plan; can {LRM}s? a preliminary evaluation of open{AI}'s o1 on
  planbench}.
\newblock In \emph{NeurIPS 2024 Workshop on Open-World Agents}.

\bibitem[{Wang et~al.(2022)Wang, Jansen, C{\^o}t{\'e}, and
  Ammanabrolu}]{wang2022scienceworld}
Ruoyao Wang, Peter Jansen, Marc-Alexandre C{\^o}t{\'e}, and Prithviraj
  Ammanabrolu. 2022.
\newblock \href {https://aclanthology.org/2022.emnlp-main.775/} {Scienceworld:
  Is your agent smarter than a 5th grader?}
\newblock \emph{arXiv preprint arXiv:2203.07540}.

\bibitem[{Wei et~al.(2022)Wei, Wang, Schuurmans, Bosma, Ichter, Xia, Chi, Le,
  and Zhou}]{DBLP:conf/nips/Wei0SBIXCLZ22}
Jason Wei, Xuezhi Wang, Dale Schuurmans, Maarten Bosma, Brian Ichter, Fei Xia,
  Ed~H. Chi, Quoc~V. Le, and Denny Zhou. 2022.
\newblock \href
  {http://papers.nips.cc/paper\_files/paper/2022/hash/9d5609613524ecf4f15af0f7b31abca4-Abstract-Conference.html}
  {Chain-of-thought prompting elicits reasoning in large language models}.
\newblock In \emph{Advances in Neural Information Processing Systems 35: Annual
  Conference on Neural Information Processing Systems 2022, NeurIPS 2022, New
  Orleans, LA, USA, November 28 - December 9, 2022}.

\bibitem[{Zhou et~al.(2024)Zhou, Xu, Zhu, Zhou, Lo, Sridhar, Cheng, Ou, Bisk,
  Fried, Alon, and Neubig}]{zhou2024webarenarealisticwebenvironment}
Shuyan Zhou, Frank~F. Xu, Hao Zhu, Xuhui Zhou, Robert Lo, Abishek Sridhar,
  Xianyi Cheng, Tianyue Ou, Yonatan Bisk, Daniel Fried, Uri Alon, and Graham
  Neubig. 2024.
\newblock \href {https://arxiv.org/abs/2307.13854} {Webarena: A realistic web
  environment for building autonomous agents}.
\newblock \emph{Preprint}, arXiv:2307.13854.

\end{thebibliography}

\appendix
\section{Additional Results}

\begin{table*}[ht]
\centering
\footnotesize
\begin{tabular}{|ll|c|c|c|c|c|c|c|}
\hline
\multicolumn{2}{|c|}{\textbf{Experiment}} & \textbf{Qwen2-72B}  & \textbf{Qwen2.5-32B} & \textbf{Qwen2.5-72B} & \textbf{LM-3.1-8B} \\ \hline
\multicolumn{1}{|l|}{\multirow{2}{*}{Basic}}
& easy & 31.2/68.7 & 56.2/93.7 & 18.7/50.0 & 37.5/62.5  \\ 
\multicolumn{1}{|l|}{}
& hard & 6.2/31.2 & 12.5/50.0 & 6.2/25.0 & 0.0/31.2  \\ \hline
\multicolumn{1}{|l|}{\multirow{2}{*}{Pre-Exploration}}
& easy & 68.7/93.7 & 56.2/81.2 & 43.7/75.0 & 50.0/100  \\ 
\multicolumn{1}{|l|}{}
& hard & 31.2/56.2 & 0.00/25.0 & 6.2/43.7 & 6.2/43.7  \\ \hline
\multicolumn{1}{|l|}{\multirow{2}{*}{Inventory Limit}}
& easy & 56.2/87.5 & 25.0/75.0 & 43.7/93.7 & 31.2/75.0  \\ 
\multicolumn{1}{|l|}{}
& hard & 25.0/56.2 &12.5/37.5 & 6.2/31.2 & 12.5/18.7 \\ \hline
\multicolumn{1}{|l|}{\multirow{3}{*}{Synthetic Words}}
& easy & 31.2/68.7 & 12.5/31.2 & 18.7/50.0 & 0.0/6.2 \\ 
\multicolumn{1}{|l|}{}
& medium & 0.0/18.7 & 0.0/6.2 & 0.0/18.7 & 0.0/0.0 \\ 
\multicolumn{1}{|l|}{}
& hard & 6.2/75.0 & 6.2/37.5 & 0.0/81.2 & 6.2/6.2 \\ \hline
\end{tabular}

\caption{Detailed results across different experiments for LLMs not shown in Table~\ref{table:detailed-experiments}. The values are \textit{Quality Score}/\textit{\% Played} for each experiment. \textit{LM-3.1-8B} $\rightarrow$ Llama-3.1-8B}
\label{tab:experiment-results-low}
\end{table*}

\begin{figure}[h]
    \centering
    \includegraphics[width=\linewidth]{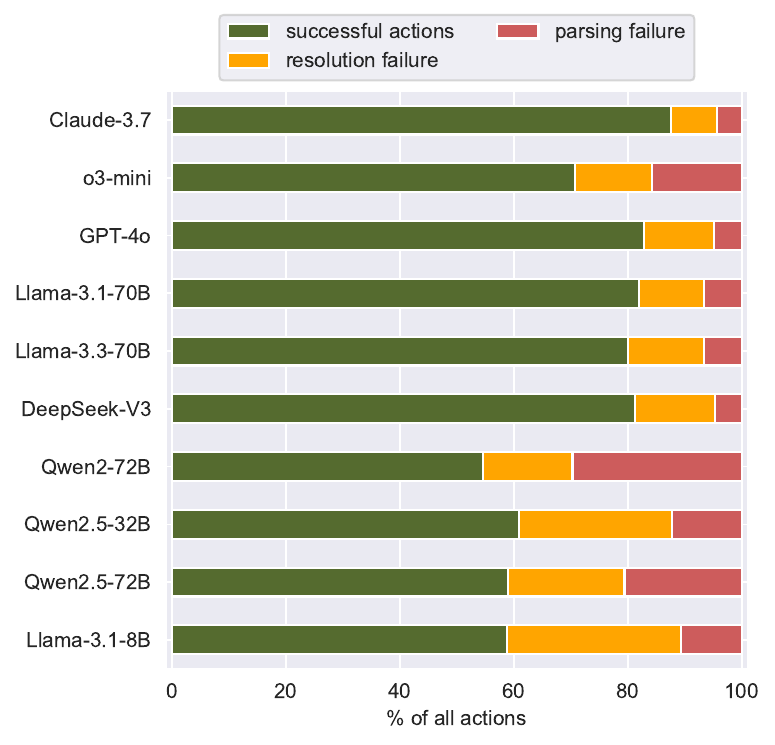}
    \caption{Successful and failed action quotas by IF interpreter processing phase.}
    \label{fig:action-phases-stack}
\end{figure}

Figure \ref{fig:action-phases-stack} shows percentages of failures by processing phase.

\begin{figure}[h]
    \centering
    \includegraphics[width=\linewidth]{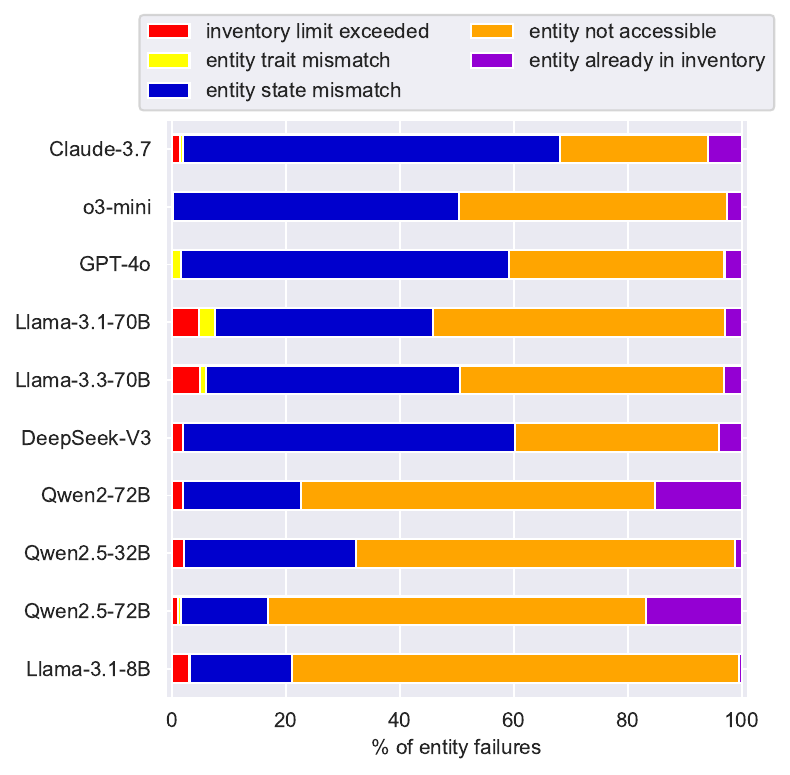}
    \caption{Percentages of selected entity-related failures for all tested models.}
    \label{fig:entity-fail-stack}
\end{figure}

Figure \ref{fig:entity-fail-stack} shows percentages of entity-related failures. o3-mini and GPT-4o do not have inventory limit failures, while Qwen2.5-32B and Qwen2.5-72B have low amounts.

\begin{figure*}[h]
    \centering
    \includegraphics[width=\linewidth]{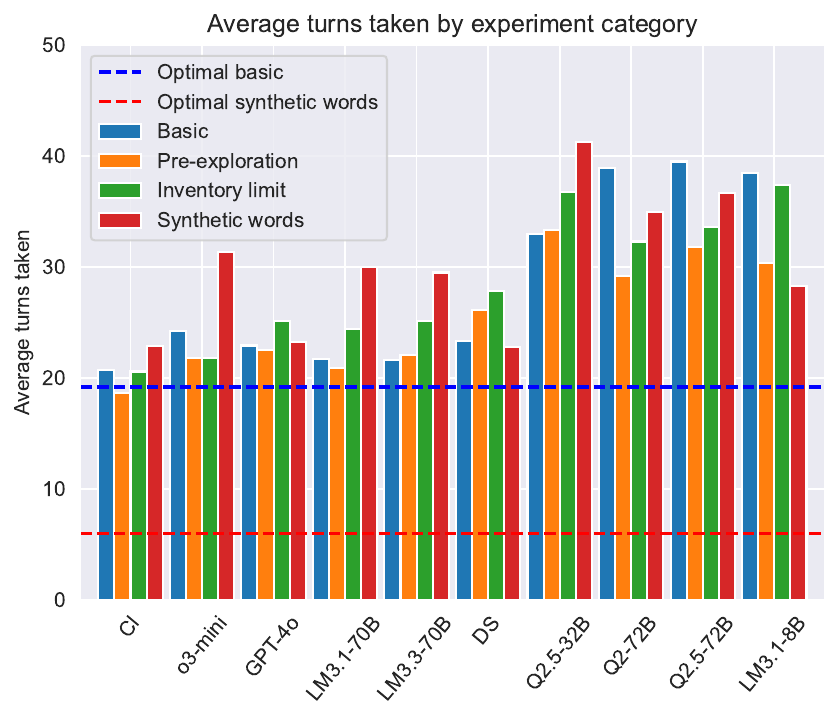}
    \caption{Average number of turns taken by each model for each experiment category. The averaged optimal number of turns for each experiment category is also given as dotted lines. Dotted blue lines for all basic, inventory limit and pre-exploration, red lines is for synthetic words experiments.}
    \label{fig:speed-plot}
\end{figure*}

\section{Human comparison results}
\begin{figure*}[h]
    \centering
    \includegraphics[width=\linewidth]{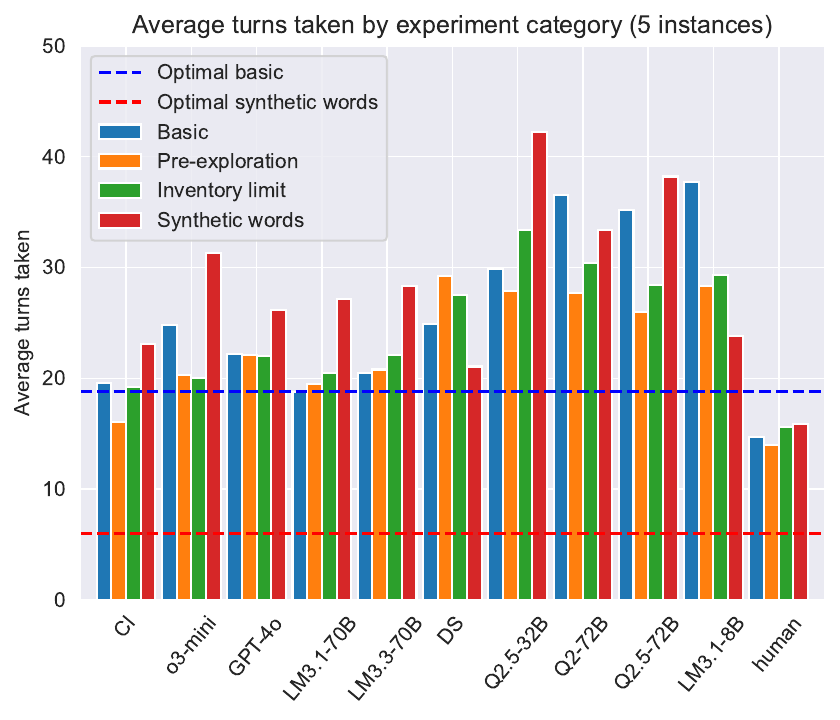}
    \caption{Average number of turns taken for the first five instances by each model and human for each experiment category. The averaged optimal number of turns for each experiment category is also given as dotted lines. Dotted blue line for basic, inventory limit and pre-exploration, red line is for synthetic words experiments.}
    \label{fig:speed-plot-human}
\end{figure*}

An annotator who has background in computational linguistics played the first five instances of the experiments, in classic IF fashion using the clemcore terminal. Table~\ref{tab:experiment-results-overview-human} shows overall scores for the tested models and human. Figure~\ref{fig:speed-plot-human} shows the average number of turns taken to solve the first five episodes by experiment category, including human data.

\begin{table*}[ht]
\centering
\footnotesize
\begin{tabular}{lrrrr}
\hline
\textbf{Model} & \makecell{\textbf{clem} \\\textbf{score}} & \makecell{\textbf{Quality} \\\textbf{Score}} & \textbf{\% Played} & \makecell{\textbf{Goal} \\\textbf{Rate}} \\
\hline
Human baseline & 100.0 & 100.0 & 100.0 & 100.0 \\ \hline
Claude-3.7 & 97.8 & 100.0 & 97.8 & 99.3 \\
o3-mini & 73.2 & 86.7 & 84.4 & 91.9 \\
GPT-4o & 51.4 & 88.9 & 57.8 & 78.5 \\
Llama-3.1-70B & 51.0 & 95.6 & 53.3 & 75.6 \\
Llama-3.3-70B & 45.6 & 93.3 & 48.9 & 71.1 \\
DeepSeek-V3 & 40.2 & 82.2 & 48.9 & 66.7 \\
Qwen2-72B & 26.3 & 62.2 & 42.2 & 54.8 \\
Qwen2.5-72B & 17.3 & 55.6 & 31.1 & 55.6 \\
Qwen2.5-32B & 14.2 & 53.3 & 26.7 & 50.4 \\
Llama-3.1-8B & 11.3 & 42.2 & 26.7 & 43.7 \\
\hline
\end{tabular}

\caption{Overview results from the first five episodes of each experiment (total 45) including human data. The annotator has a background in computational linguistics and participated in the study voluntarily.}
\label{tab:experiment-results-overview-human}
\end{table*}

\section{IF Actions, Objects \& Rooms} \label{sec:basic-domain}
All experiment instances have the core entities ``player'' and ``inventory'' needed for the core AdventureGame IF interpreter. These and the 'floor' entity can not be replaced for the synthetic words \textit{easy} and \textit{medium} experiments.

The basic home domain has the following entities (with possible location rooms in brackets): Table (kitchen, living room), side table (living room, bedroom), counter (kitchen), refrigerator (kitchen, pantry), cupboard (kitchen), wardrobe (bedroom), shelf (kitchen, pantry, living room), freezer (pantry), potted plant (living room, hallway, bedroom), chair (living room), bed (bedroom), couch (living room), broom (broom closet), mop (broom closet), sandwich (kitchen, pantry), apple (kitchen, pantry), banana (kitchen, pantry), orange (kitchen, pantry), peach (kitchen, pantry), plate (kitchen), book (living room, bedroom), pillow (bedroom).

The following entities are ``supports'', allowing ``movable'' objects to be placed ``on'' them: Table, side table, counter, shelf, chair, bed, couch. The following entities are 'containers', allowing ``movable'' objects to be placed ``in'' them and objects in them being accessible if they are 'open': Refrigerator, cupboard, wardrobe, freezer.

The following entities are ``movable'': Potted plant, broom, mop, sandwich, apple, banana, peach, plate, book, pillow.

The basic home domain has the following rooms (with possible adjacent rooms in brackets): Kitchen (pantry, living room, hallway), pantry (kitchen, hallway), hallway (kitchen, pantry, living room, broom closet), living room (kitchen, hallway), broom closet (hallway) and bedroom (living room, hallway). All can be replaced for the synthetic words \textit{easy} and \textit{medium} experiments.

Table \ref{tab:actions} lists the actions defined for all basic experiments, including which can be replaced for the synthetic words \textit{easy} and \textit{medium} experiments.

\begin{table*}
\footnotesize
    \centering
    \begin{tabular}{llllll}
         Action & Targets & Description & Epistemic & Pragmatic & Replaceable \\
         \hline
         \texttt{open} & ``container'' entities & Changes state of \texttt{closed} & Yes & Yes & Yes \\
         & & container entity to \texttt{open} & & & \\
         \hline
         \texttt{close} & ``container'' entities & Changes state of \texttt{open} & No & Yes & Yes \\
         & & container entity to \texttt{closed} & & & \\
         \hline
         \texttt{take} & ``takeable'' entities & Removes \texttt{in/on} state for & No & Yes & Yes\\
         & & ``takeable'' entity and adds & & \\
         &&\texttt{in(entity,inventory)} fact & & \\
         \hline
         \texttt{put}&``takeable''\& & Removes \texttt{in(entity,inventory)} & No & Yes & Yes\\
         &``container''/``support''& state for ``takeable'' entity and & & \\
         & entities & adds \texttt{in/on(entity,target)} fact & & \\
        \hline
         \texttt{go}& ``room'' & Changes \texttt{at} state of player entity & Yes & Yes & No\\
         &&and all entities in inventory & &\\
         && to target room & & \\
         \hline
         \texttt{done}& - & Ends the episode & No & Yes & No\\
         \hline
         \texttt{examine}& entities & Results in entity state feedback & Yes & No & No \\
         \hline
         \texttt{look}& 'room' & Results in current room & No & No & No \\
         &  & description feedback &  &  & 
    \end{tabular}
    \caption{Basic action types used in AdentureGame. Targets are those for which the world state holds a fact assigning the listed state. Replaceable denotes actions that can be replaced with synthetic words in the \textit{easy} and \textit{medium} synthetic words experiments.}
    \label{tab:actions}
\end{table*}

\section{Initial Prompts} \label{sec:initial-prompts}
Prompt template for 'basic' and pre-exploration variant instances:
\begin{verbatim}
You are playing a text adventure game. I 
will describe what you can perceive in 
the game. You write the single action you 
want to take in the game starting with >. 
Only reply with an action.
For example:
> open cupboard

Your goal for this game is: $GOAL$

Once you have achieved your goal, write 
"> done" to end the game.

\end{verbatim}

\noindent Prompt template for limited inventory variant instances:

\begin{verbatim}
You are playing a text adventure game. I 
will describe what you can perceive in 
the game. You write the single action you 
want to take in the game starting with >. 
Only reply with an action.
For example:
> open cupboard

You can have up to two objects in your 
inventory at the same time.

Your goal for this game is: $GOAL$

Once you have achieved your goal, write 
"> done" to end the game.

\end{verbatim}

\noindent Prompt template for synthetic word variant instances:

\begin{verbatim}
You are playing a text adventure game. I 
will describe what you can perceive in 
the game. You write the single action you 
want to take in the game starting with >. 
Only reply with an action.
For example:
> open cupboard

$NEW_WORDS_EXPLANATIONS$

Your goal for this game is: $GOAL$

Once you have achieved your goal, write 
"> done" to end the game.
\end{verbatim}

\noindent A description of the room the player starts in is appended to complete the initial prompts.

\noindent The placeholder \texttt{\$GOAL\$} in the templates is replaced with the task goal.

Example ``easy'' difficulty task goal: \texttt{Put the pillow on the table, the book on the table and the plate on the table.}

Example ``hard'' difficulty task goal: \texttt{Put the pillow on the counter, the book on the shelf and the plate on the table.}

\noindent For the synthetic words experiments, \texttt{\$NEW\_WORDS\_EXPLANATIONS\$} is replaced with \texttt{In addition to common actions, you can } followed by a synthetic action word and its explanation for \textit{easy}, or a list of synthetic actions words for \textit{medium} and \textit{hard} instances.

\section{Environment Graphs}
\begin{figure*}[ht]
    \centering
    \includegraphics[width=\linewidth]{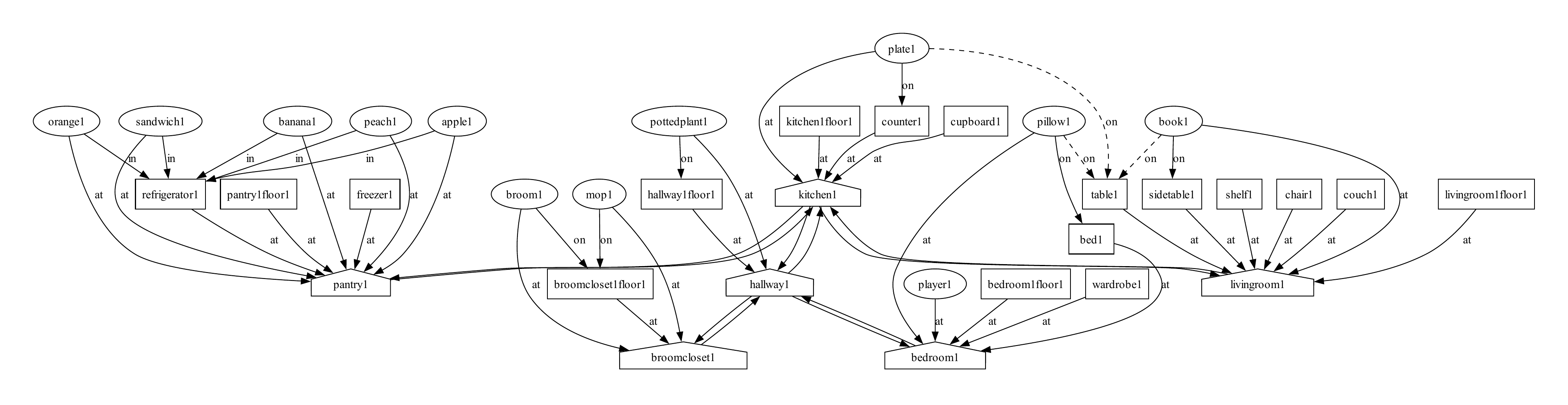}
    \caption{Graph representation of an ``easy'' basic instance.}
    \label{fig:easy-instance-graph}
\end{figure*}

\begin{figure*}[ht]
    \centering
    \includegraphics[width=\linewidth]{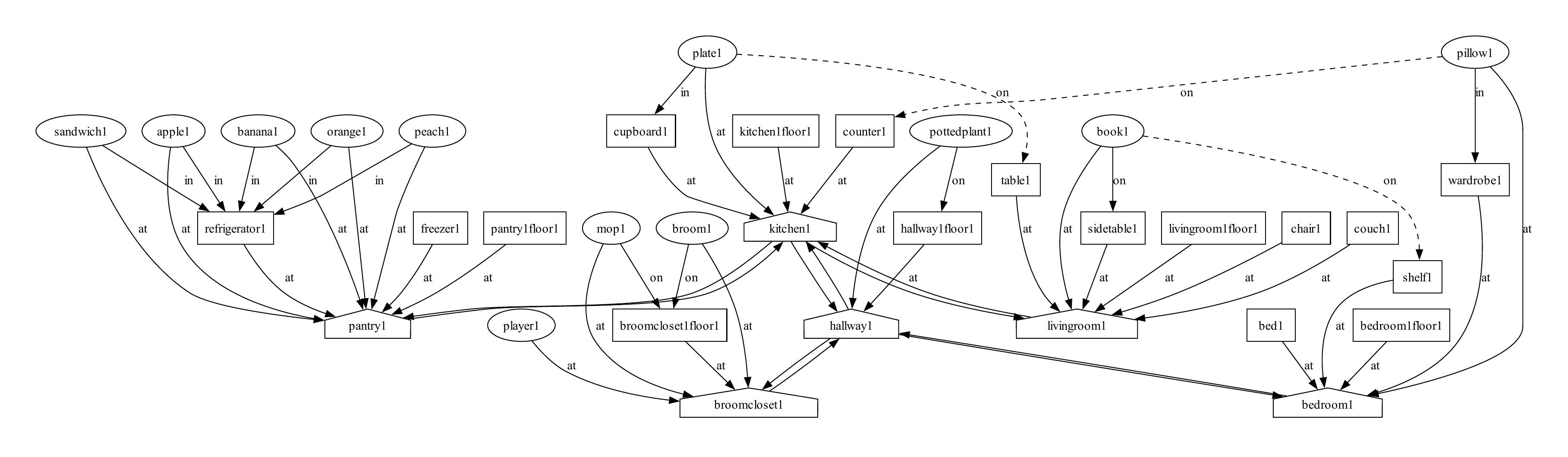}
    \caption{Graph representation of a ``hard'' basic instance.}
    \label{fig:hard-instance-graph}
\end{figure*}

To illustrate differences between ``easy''/``hard'' environment and task complexity, Figures \ref{fig:easy-instance-graph} and \ref{fig:hard-instance-graph} show graph representations of initial game world states and task targets. House-shaped nodes are rooms, with arrow edges showing bidirectional connections between them. Round nodes are ``movable'' entities, connected to rectangular receptacles and rooms by edges labelled with their prepositional state. Dashed edges connect the movable task objects to their target receptacles and are labelled with the target prepositional state.

\end{document}